\documentclass[letterpaper, 10 pt, conference]{ieeeconf}  

\IEEEoverridecommandlockouts                              

\usepackage{graphics} 
\usepackage{epsfig} 
\usepackage{amsmath} 
\usepackage{amssymb}  
\usepackage{algorithmic}
\usepackage{algorithm}

\newcommand{\q}[1]{``#1''}

\title{\LARGE \bf
Modeling neural dynamics during speech production \\
using a state space variational autoencoder
}
\author{Pengfei Sun, David A.\ Moses, and Edward F.\ Chang%
\thanks{P.\ Sun, D.\ Moses, and E.\ Chang are with the Department of Neurological Surgery and Center for Integrative Neuroscience, UC San Francisco, CA, USA. This work was funded by a research contract under Facebook's Sponsored Academic Research Agreement.
}%
}

\begin{document}

\maketitle
\thispagestyle{empty}
\pagestyle{empty}

\begin{abstract}
Characterizing the neural encoding of behavior remains a challenging task in many research areas due in part to complex and noisy spatiotemporal dynamics of evoked brain activity.
An important aspect of modeling these neural encodings involves separation of robust, behaviorally relevant signals from background activity, which often contains signals from irrelevant brain processes and decaying information from previous behavioral events.
To achieve this separation, we develop a two-branch State Space Variational AutoEncoder (SSVAE) model to individually describe the instantaneous evoked foreground signals and the context-dependent background signals.
We modeled the spontaneous speech-evoked brain dynamics using smoothed Gaussian mixture models.
By applying the proposed SSVAE model to track ECoG dynamics in one participant over multiple hours, we find that the model can predict speech-related dynamics more accurately than other latent factor inference algorithms.
Our results demonstrate that separately modeling the instantaneous speech-evoked and slow context-dependent brain dynamics can enhance tracking performance, which has important implications for the development of advanced neural encoding and decoding models in various neuroscience sub-disciplines.
\end{abstract}

\section{INTRODUCTION}

Modeling behaviorally relevant brain activity is an important aspect of many fields of computational neuroscience.
Typically, when recording neural activity during a behavioral task, the recorded signal is a combination of the evoked activity of interest, which is immediately relevant to ongoing behavior, and signals from various other sources of \q{noise}.
Although some types of noise are relatively predictable and manageable (such as line noise), it is difficult to model signal contributions from behaviorally irrelevant background processing and residual processing from recent behavioral events.
Both the signal of interest and the noise component can have non-stationary and time-variant dynamics that depend on the behavioral state of the subject, further contributing to this difficulty.
Advanced modeling techniques capable of decoupling these two components could have major implications in neural time series analyses, effectively increasing the signal-to-noise ratio in recorded data through computational approaches alone.

State space modeling (SSM) as a time domain approach has been widely used in modeling dynamical systems and generalizes other popular time series models \cite{anderson2011statistical}.
In brain signal modeling approaches, conventional linear time-invariant SSM models (e.g., Kalman filters) have been used to predict arm movement trajectories from monkey's spike recordings \cite{makin2018superior}.
In these models, the transition matrices are provided as prior knowledge.
Shanechi developed a closed-loop SSM \cite{shanechi2016robust}, focusing on subspace identification to estimate time-variant SSMs online.
A different type of dynamical system introduced by Yang uses parameter matrices that update over time, proving effective in tracking ECoG signals \cite{yang2017dynamic}. 

Recent developments in unsupervised learning use a variational Bayesian approach to implement SSM estimation and inference \cite{sussillo2016lfads}.
Frigola proposed a framework focusing on Bayesian learning of non-parametric nonlinear SSMs \cite{frigola2014variational}.
By using sparse Gaussian processes to encode dynamical systems, their variational training procedure enables learning of complex systems without risk of overfitting.
Karl \textit{et al.} utilize stochastic gradient variational Bayes modeling as the inference mechanism for highly nonlinear SSMs \cite{karl2016deep}.
Fraccaro introduced the Kalman Variational Autoencoder (KVAE) to disentangle temporal sequences in a latent space that describes nonlinear dynamics \cite{fraccaro2017disentangled}.
Sussillo \textit{et al.} applied Variational Autoencoder (VAE) and recurrent neural network (RNN) models as the generative and state transition framework to successfully track neural activity over time \cite{sussillo2016lfads}.
In Krishnan's work, nonlinear SSMs are parameterized by RNNs, allowing the inference and generative models to be learned simultaneously \cite{krishnan2017structured}.
Although these approaches have been successful in modeling various types of signals, they might not be suitable for modeling highly non-stationary neural processes that often operate on multiple timescales.
Ideally, instantaneous and behaviorally-relevant neural dynamics could be separated from slow, context-dependent background processes to model the relationship between neural activity and rapidly changing behavior, such as during speech production tasks.

In this study, we develop a general State Space Variational Autoencoder (SSVAE) model to track speech-related ECoG signals.
One human participant verbally produced speech sentences while we simultaneously recorded ECoG signals from a high-density electrode array.
We then applied our SSVAE model to separate the recorded neural activity into the relevant signal evoked by instantaneous speech (currently occurring speech behavior) and the background signal (noise and residual signal from preceding behavior).
Compared with SSMs based on variational Gaussian processes \cite{fraccaro2017disentangled,frigola2014variational}, our model generalizes the inputs (in other words, inducing variables) as unknown components.
We then examine the non-stationary nature of the high-density ($256$ electrodes) ECoG data by comparing the predictions of the proposed SSVAE with other latent dynamics inference approaches.
Using the separated components of the ECoG signals, we find that the proposed SSVAE model can reconstruct the original signals more accurately than other sequential models.
These findings indicate that the relevant instantaneous signal can be reliably extracted from the neural activity, suggesting that the SSVAE model could be used as a latent feature extraction approach for decoding speech (and potentially other types of behavior) from neural activity.

\section{METHODOLOGY}
\subsection{Background}
\begin{figure}
  \centering
  \includegraphics[scale=0.6]{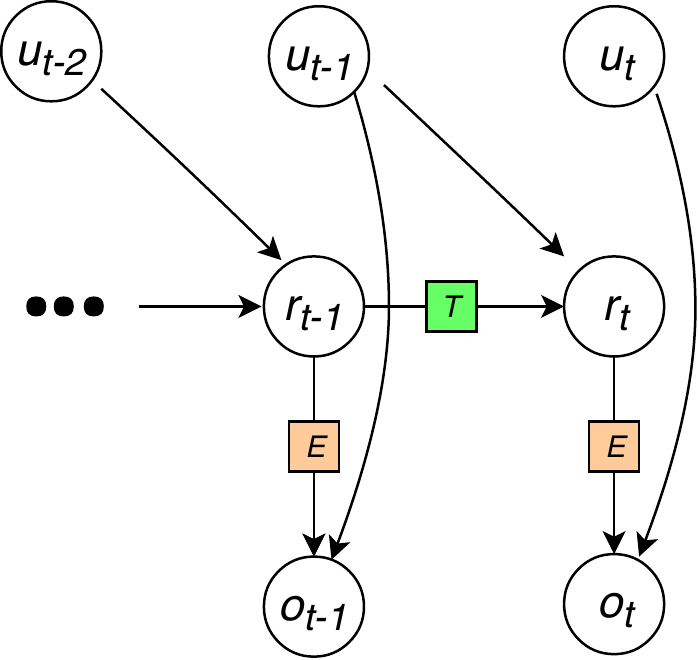}
  \caption{Generative model for sequential data}
  \label{generative}
  \vspace{-4mm}
\end{figure}

\paragraph{State Space models} The general SSM structure in $n$-dimensional data space can be described as:
\begin{equation}
\begin{split}
    \mathbf{r}_{t+1} &= f_{t}(\mathbf{r}_{t}, \mathbf{u}_{t}, \mathbf{v}_{t}, \theta) \\
    \mathbf{o}_{t}   &= h_{t}(\mathbf{r}_{t}, \mathbf{u}_{t}, \mathbf{e}_{t}, \theta) .
\end{split} 
\label{SSM}
\end{equation}
Here, $\mathbf{r}_{t} \in \mathcal{R}^{r_{1}\times \ldots \times r_{n}}$ denotes the state variable, with variables $\mathbf{u}_{t} \in \mathcal{R}^{u_{1}\times \ldots \times u_{n}}$ and $\mathbf{o}_{t} \in \mathcal{R}^{o_{1}\times \ldots \times o_{n}}$ denoting inputs and observations, respectively.
Typically, the $\mathbf{u}_{t}$ values are observable, but in the SSVAE implementation they are hidden.
$\mathbf{v}_{t}$ and $\mathbf{e}_{t}$ represent mutually independent random components, and $\theta \in \mathcal{R}^{l_{1}^{\theta}\times \ldots \times l_{n}^{\theta}}$ represents unknown parameters that specify the mappings $f_{t}$ and $h_{t}$, which may be nonlinear and time-varying \cite{schon2011system}.
Here, we assume that the distribution of the latent states and inputs are multivariate Gaussians with means and covariances that are differentiable functions of previous latent states and inputs.
Eq. (\ref{SSM}) subsumes a large family of linear and nonlinear SSMs.
By making explicit assumptions about the unknown systems $f_{t}$ and $h_{t}$, we obtain a separable form of the linear Gaussian SSM described by:
\begin{equation}
\begin{split}
    \mathbf{r}_{t+1} = & f_{t}^{1}(\mathbf{r}_{t}, \theta)\mathbf{r}_{t} + f_{t}^{2}(\mathbf{u}_{t}, \theta)\mathbf{u}_{t} + \mathbf{v}_{t} \\
    \mathbf{o}_{t}   = & h_{t}^{1}(\mathbf{r}_{t}, \theta)\mathbf{r}_{t} + h_{t}^{2}(\mathbf{u}_{t}, \theta)\mathbf{u}_{t} + \mathbf{e}_{t} ,
\end{split}
\label{LSSM}
\end{equation}
where the implicit functions ($f_{t}$ and $h_{t}$) and variables ($\mathbf{v}_{t}$ and $\mathbf{e}_{t}$) in Eq. (\ref{SSM}) are processed as independent components (e.g., $f_{t}^{1}$ and  $h_{t}^{1}$).
Equivalently, the emission and state transition models can be described as:
\begin{equation}
    p_{\theta}(\mathbf{o}_{1:T}|\mathbf{r}_{1:T}, \mathbf{u}_{1:T}) = \prod_{t=1}^{T}p_{\theta}(\mathbf{o}_{t}|\mathbf{r}_{t}, \mathbf{u}_{t}),
    \label{emission}
\end{equation}
\begin{equation}
    p_{\eta}(\mathbf{r}_{1:T}| \mathbf{u}_{1:T}) = \prod_{t=1}^{T-1}p_{\eta}(\mathbf{r}_{t+1}|\mathbf{r}_{t}, \mathbf{u}_{t}).
    \label{transition}
\end{equation}
Eqs. (\ref{emission}) and (\ref{transition}) both assume that the hidden state $\mathbf{r}_{t}$ contains all necessary information about the current observation $\mathbf{o}_{t}$ as well as the next state $\mathbf{r}_{t+1}$ (given the current control input $\mathbf{u}_{t}$ and transition and emission parameters $\eta_{t}$ and $\theta$), and the details about transition $\tau$ and emission $E$  are illustrated in Fig. \ref{generative}.
  
\paragraph{Variational autoencoder}
A variational autoencoder (VAE) defines a deep generative model $p(\mathbf{o}_{t}, \mathbf{r}_{t}, \mathbf{u}_{t})$ $=$ $p(\mathbf{o}_{t}|\mathbf{r}_{t}, \mathbf{u}_{t})p(\mathbf{r}_{t}, \mathbf{u}_{t})$ for observations $\mathbf{o}_{t}$ by introducing latent encodings $\mathbf{r}_{t}$ and $\mathbf{u}_{t}$.
Given a likelihood $p(\mathbf{o}_{t}|\mathbf{r}_{t}, \mathbf{u}_{t})$ and a prior $p(\mathbf{r}_{t}, \mathbf{u}_{t})$, the posterior $p(\mathbf{r}_{t}, \mathbf{u}_{t}|\mathbf{o}_{t})$ represents a stochastic map from $\mathbf{o}_{t}$ to the manifold of $\mathbf{r}_{t}$ and $\mathbf{u}_{t}$.
As this posterior is often analytically intractable, VAEs approximate it with a variational distribution $q_{\phi}(\mathbf{r}_{t}, \mathbf{u}_{t}|\mathbf{o}_{t})$ that is parameterized  by $\phi$.
The approximation $q_{\phi}$ is commonly called the inference or encoding network. 

\subsection{State Space Variational AutoEncoder}
Using Eq. (\ref{SSM}) to represent our experimental task, the observed speech-related ECoG signals $\mathbf{o}_{t}$ consist of three components: Non-speech brain activities and general recording noise $\mathbf{e}_{t}$, previous speech-evoked brain dynamics $h_{t}^{1}\mathbf{r}_{t}$, and instantaneous speech-evoked brain dynamics $h_{t}^{2}\mathbf{u}_{t}$.
By isolating the instantaneous speech responses, these features can be used to learn reliable mappings between the neural activity and the speech behavior for encoding and decoding purposes.
Considering both $h_{t}$ and $f_{t}$ are system functions and are generally independent from speech behavior, the majority of speech-evoked brain dynamics can be represented by the latent features $\mathbf{r}_{t}$ and $\mathbf{u}_{t}$.

The \emph{State Space Variational Autoencoder} (SSVAE) model is based on the concept described above.
Each observation frame $\mathbf{o}_{t}$ is encoded into two latent feature points ($\mathbf{r}_{t}$ and $\mathbf{u}_{t}$) in a low-dimensional manifold.
Both latent features are used as the hidden states and inputs in the transition function of the SSM. 

To model the emission function, a VAE decoder is applied to generate the output pseudo-observation $\hat{\mathbf{o}}_{t}$.
For each input segment $\mathbf{o}_{1:T}$, the SSVAE contains $T$ separate VAEs that share the same decoder $p_{\theta}(\mathbf{o}_{t}|\mathbf{r}_{t},\mathbf{u}_{t})$ and encoder $q_{\phi}(\mathbf{r}_{t},\mathbf{u}_{t}|\mathbf{o}_{t})$.
These models depend on each other through a time-dependent prior over $\mathbf{r}$.
\begin{figure}
    \centering
    \includegraphics[scale=0.33]{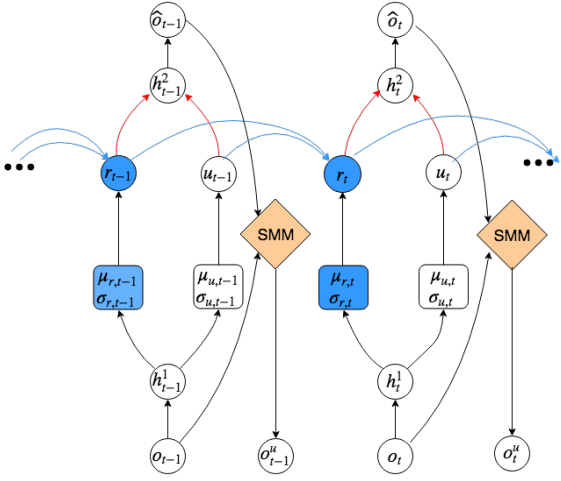}
    \caption{State Space based Auto Encoder}
    \label{udssm}
    \vspace{-4mm}
\end{figure}

Here, our VAEs incorporate deep neural networks to tackle the difficult latent variable inference problem, where exact model inference is intractable and conventional approximate methods do not scale well.
Theoretically, a VAE can be an approximated using variational Bayesian estimation.
Compared with KVAEs \cite{fraccaro2017disentangled}, the proposed SSVAE provides a two-branch variational encoding, which assumes the input components $\mathbf{u}_{t}$ are unknown.
The proposed SSVAE model is effectively a generalized extension of KVAE.
To obtain the prior of $\mathbf{u}_{t}$, unsupervised Smoothed Gaussian Mixture Model (SMM) is incorporated to estimate the instantaneous components $\mathbf{o}_{t}^{u} = h_{t}^{2}(\mathbf{u}_{t}, \gamma)\mathbf{u}_{t}$ as illustrated in Fig. \ref{udssm}.

\paragraph{Generative model} 
The output of the autoencoder $\hat{\mathbf{o}}_{t}$ is generated from the latent representation $\mathbf{r}_{t}$ and $\mathbf{u}_{t}$.
The generative model is described as an underlying latent dynamical system with emission model $p(\mathbf{o}_{1:T}|\mathbf{r}_{1:T}, \mathbf{u}_{1:T})$ and transition model $p(\mathbf{r}_{1:T}|\mathbf{u}_{1:T})$:
\begin{equation}
\begin{split}
    p(\mathbf{o}_{1:T}|\mathbf{u}_{1:T}) = \int & p_{\theta}(\mathbf{o}_{1:T}|\mathbf{r}_{1:T}, \mathbf{u}_{1:T}) \\
    & p_{\eta}(\mathbf{r}_{1:T}|\mathbf{u}_{1:T}) d\mathbf{r}_{1:T} ,
\end{split}
\label{likelihood}
\end{equation}
where $\mathbf{r}_{1:T}$ $\left( \mathbf{r}_{t} \in \mathcal{R}^{l_{1}^{r}\times \ldots \times l_{n}^{r}} \right)$ denotes the corresponding latent sequence.
Efficient posterior inference distributions $p(\mathbf{r}_{1:T}, \mathbf{u}_{1:T}| \mathbf{o}_{1:T})$ are required to use the system effectively.
In this paper, $p(\mathbf{r}_{t}, \mathbf{u}_{t}|\mathbf{o}_{t})$ is modeled by the autoencoder inference network (parameterized by $\phi$).

\paragraph{Learning and inference for the SSVAE}
We learn $\theta$ and $\eta$ from a set of observations ${\mathbf{o}_{1:T}}$ by maximizing the sum of their respective log likelihoods $\sum_{T}\log p_{\theta\eta}(\mathbf{o}_{t})$ as a function of $\theta$ and $\eta$.
For notational simplicity, we drop the index.
The log likelihood or evidence is an intractable average over all plausible $\mathbf{r}$ and $\mathbf{u}$.
A tractable approach to both learning and inference is to introduce a variational distribution $q_{\phi}(\mathbf{r}, \mathbf{u}|\mathbf{o})$ that approximates the posterior.
The evidence lower bound (ELBO) $\mathcal{L}$ is
\begin{equation}
\begin{split}
    \log p(\mathbf{o})  &= log \int q_{\phi}(\mathbf{r}, \mathbf{u}|\mathbf{o}) \frac{p(\mathbf{o}, \mathbf{r}, \mathbf{u})}{q_{\phi}(\mathbf{r}, \mathbf{u}|\mathbf{o})} \\
   &\geq \int q_{\phi}(\mathbf{r}, \mathbf{u}|\mathbf{o}) \log \frac{p(\mathbf{o}, \mathbf{r}, \mathbf{u})}{q_{\phi}(\mathbf{r}, \mathbf{u}|\mathbf{o})}  \\
  = &\mathbb{E}_{q_{\phi}(\mathbf{r}, \mathbf{u}|\mathbf{o})}\bigg[\log \frac{p_{\theta}(\mathbf{o}|\mathbf{r}, \mathbf{u})p_{\eta}(\mathbf{r}|\mathbf{u})p_{\eta}(\mathbf{u})}{q_{\phi}(\mathbf{r}, \mathbf{u}|\mathbf{o})} \bigg] = \mathcal{L},
\end{split}
\end{equation}
and a sum of $\mathcal{L}(\theta, \eta, \phi)$ is maximized instead of a sum of log likelihoods.
The variational distribution $q$ depends on $\phi$, but for a tight bound we specify $q$ to be equal to the posterior distribution that only depends on $\theta$ and $\eta$.
Compared to previous variational inference and state space based generative models, the proposed SSVAE model regards the input components $\mathbf{u}$ as unknown.
We structure $q$ so that it incorporates the exact conditional posterior $p_{\eta}(\mathbf{r}|\mathbf{u})$, which we obtain from the state space transition equation, as a factor of $\eta$:
\begin{equation}
\begin{split}
     q(\mathbf{r}, \mathbf{u} | \mathbf{o}) & = q_{\phi}(\mathbf{u}|\mathbf{o}) p_{\eta}(\mathbf{r}|\mathbf{u}) \\
     & = \prod_{t=1}^{T}q_{\phi}(\mathbf{u}_{t}|\mathbf{o}_{t})p_{\eta}(\mathbf{r}|\mathbf{u}).
\end{split}
\end{equation}
In this work, $q_{\phi}(\mathbf{u}|\mathbf{o})$ is the encoding network that maps $\mathbf{o}$ to the instantaneous feature space.
$p_{\eta}(\mathbf{r}|\mathbf{u})$ represents the transition function of the SSM.
Accordingly, the ELBO can be written as:
\begin{equation}
\begin{split}
    \mathcal{L}(\theta, \eta, \phi) = &\mathbb{E}_{q_{\phi}(\mathbf{r}, \mathbf{u}|\mathbf{o})}[ \log p_{\theta}(\mathbf{o}|\mathbf{r}, \mathbf{u})] \\
    -&\mathbb{E}_{q_{\phi}(\mathbf{r}, \mathbf{u}|\mathbf{o})}[\log p_{\eta}(\mathbf{r}|\mathbf{u})] \\
    -&\mathcal{D}(q_{\phi}(\mathbf{u}|\mathbf{o})\|p_{\eta}(\mathbf{u})).
\end{split}
\label{ELBO}
\end{equation}
In Eq. (\ref{ELBO}), the first component represents the expectation of reconstructed observation, the second component refers to the expectation of hidden state conditioned on instantaneous features $u_{t}$ and the $\mathcal{D}$ in the third component is Kullback-Leibler (KL) divergence. $q_{\phi}(\mathbf{u}|\mathbf{o})$ can be inferred by the encoding network $q_{\phi}$, but inferring the instantaneous features $p(\mathbf{u})$ is intractable with unknown inputs. 

To obtain priors $p_{\eta}(\mathbf{u})$, a smoothed GMM (SMM) model is applied to estimate the unknown input components $\mathbf{o}_{u}$.
By feeding $\mathbf{o}_{u}$ into the encoding network $q_{\phi}$, we infer $p(\mathbf{u}^{\prime})$ as an approximation to $p_{\eta}(\mathbf{u})$.
This approximation assumes that the two branches of the inference network $q_{\phi}$ provide orthogonal projections.
To constrain the estimation of $p_{\eta}(\mathbf{u})$ and avoid the degradation of the SMM into \q{model collapse}, a combination of inputs and predictions are used to estimate the mixture components.
To constrain the pre-estimation and avoid the SMM degraded to Maximum Likelihood driven model, a combination of inputs $o_{t}$ and predictions $\hat{o}_{t}$ are used to implement the mixture components estimation: 
\begin{equation}
    \mathbf{o}_{t}^{\prime} = \gamma \mathbf{o}_{t} + (1-\gamma)\mathbf{\hat{o}}_{t},   
\end{equation}
where $\gamma$ is the weighting coefficients. 

By using $p(\mathbf{u}^{\prime})$ to replace $p_{\eta}(\mathbf{u})$, the transition network parameterized by $\eta$ is applied to infer the $q_{\eta}(\mathbf{r}|\mathbf{u})$ in Eq. (\ref{ELBO}). Accordingly, the loss function of the model is written as:
\begin{equation}
\begin{split}
\mathcal{L}_{c} &= \alpha_{1} \| \hat{o}_{t}-o_{t}\|_{F}^{2} \\
& + \alpha_{2}\mathcal{D}((\mu_{t, r}, \sigma_{t, r})\|(\mu_{t-1, r}, \sigma_{t-1, r})) \\
& + \alpha_{3} \mathcal{D}((\mu_{t, u}, \sigma_{t, u})\|(\mu_{t, u}^{\prime}, \sigma_{t, u}^{\prime})).
\end{split}
\label{obj}
\end{equation}
In Eq. (\ref{obj}), the first KL divergence term in the loss function encourages the transition model to pull $p(\mathbf{r}_{t})$ to match $p(\mathbf{r}_{t-1})$, and the second KL divergence term pulls $q_{\phi}(\mathbf{u}|\mathbf{o})$ to match $p(\mathbf{u}^{\prime})$.  



\subsection{Model evaluation}
We divided the raw ECoG signals recorded from a single participant $O_{t}$ (sampled at approximately $95$ Hz) into training and testing datasets.
Using the training dataset, we separately train long short-term memory VAE (LSTM-VAE), vanilla-VAE, unscented Kalman filter (UKF), and proposed SSVAE models.
The comparisons of the four models with respect to tracking ECoG dynamics and speech detection are implemented on the testing dataset. 
Because the instantaneous features $u_{t}$ should reflect the spontaneous speech-evoked ECoG components, a simple analysis to assess their efficacy was to use them to detect which time points occurred while the participant was speaking.
The total duration of produced speech in the training and testing blocks was $45$ and $30$ minutes, respectively, and the time window T is 6 seconds.

To compare the four models, we investigated their performance with respect to dynamics prediction and speech detection.
For dynamics prediction, the relative prediction error (RPE) is defined as:
\begin{equation}
    RPE = \frac{1}{n_{c}}\sum_{i=1}^{n_{c}}\sqrt{\frac{\sum_{t=1}^{T}(\hat{o}_{t}^{i}-o_{t}^{i})^{2}}{\sum_{t=1}^{T}(o_{t}^{i})^{2}}},
\label{RPE}
\end{equation}
where $T$ is the total number of time samples in the test dataset and the upper index $i$ denotes the electrode channel number.
For speech detection accuracy, the commonly used speech hit rate (SHR) metric is used.
To obtain binary classifications (between speech and silence), the amplitude power of $\hat{u}_{t}$ for the SSVAE model and the hidden state power for the other three algorithms are normalized into the range $\left[ -1, 1 \right]$.
A threshold value of $0$ is then used to classify each time point.
For the SSVAE model, this is defined as:    
\begin{equation}
    \hat{u}_{t} = \mathcal{F}\Bigg(\sum_{i=1}^{n_{c}}u_{t, i}^{2} \Bigg), 
\end{equation}
where $\mathcal{F}$ is a low pass filter and $i$ indexes the channels.

\section{RESULTS}
\subsection{Dynamics tracking}
\begin{figure}[ht]
    \centering
    \includegraphics[scale=0.5]{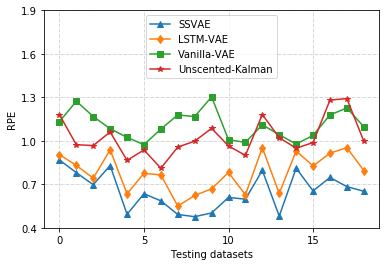}
    \vspace{-2mm}
    \caption{Reconstruction errors on testing datasets (on a single participant across 19 blocks). Lower values indicate better performance.}
    \label{ecogrecon}
\end{figure}

By comparing the reconstructed amplitudes of the raw ECoG signals obtained by four algorithms in Fig. \ref{ecogrecon}, we show that our SSVAE model outperforms the other models at tracking the neural dynamics, achieving the lowest RPE values in each of the 19 testing blocks.
As defined in Eq. \ref{RPE}, an RPE of $1$ indicates a mean prediction (i.e., reconstructing the observation value simply as its mean, which is $0$ here) and provides no real tracking power.
The SSVAE model achieves an RPE value less than $1$ for each test block, demonstrating its capability to consistently track the ECoG signals.

\subsection{Speech Detection}
\begin{figure}
    \centering
    \includegraphics[scale=0.5]{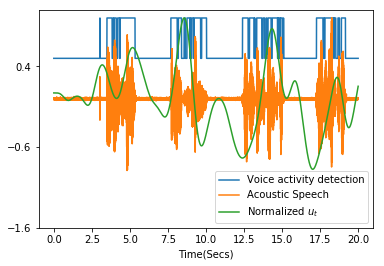}
    \vspace{-2mm}
    \caption{Instantaneous feature $u_{t}$ (green), acoustic speech signal (orange) and the binary voice activity detection results in time domain. The amplitudes of $u_{t}$ are normalized to fit the overlap plotting. Low pass filtering is applied to smooth $u_{t}$.}
    \label{speechdetection}
    \vspace{-2mm}
\end{figure}

In Fig. \ref{speechdetection}, the notched binary lines indicate the voice activities.
High amplitudes in the acoustic speech signal signify the occurrence of speech production, and low amplitudes signify silence.
The normalized instantaneous feature component $\hat{u}_{t}$ is active during the major speech segments.
Due to the strong constraint imposed by SMM, the inference of spontaneous features may temporally deviate from the speech activities.
This is visible as a delayed response to speech production between the $15$- and $17.5$-second marks.
\begin{table}[!ht]
    \centering
    \caption{Speech detection accuracy by four approaches}
    \begin{tabular}{c  c  c  c  }
    \hline
     SSVAE         & LSTM$-$VAE      &Vanilla$-$VAE    & Unscented$-$Kalman \\ \hline 
     0.51 $\pm$ 0.11  & 0.42$\pm$ 0.13  & 0.29$\pm$ 0.05   & 0.37 $\pm$ 0.06     \\  \hline
    \end{tabular}
    \label{tab:sda}
\end{table}
Table \ref{tab:sda} summarizes the average speech detection accuracies obtained by the four models across the 19 testing blocks.
The SSVAE performance is slightly better than that of the LSTM-VAE approach and much better than the UKF and vanilla-VAE approaches.

\section{CONCLUSIONS}
In this study, we propose a state space VAE model to obtain instantaneous speech-evoked ECoG features. This two-branch latent inference approach can effectively track the dynamics of the neural signals.
By introducing the instantaneous speech-evoked ECoG components as unknown inputs, the SSVAE model attempts to maximize the likelihood of the observations (i.e., raw ECoG) by jointly inferring the residuals (hidden states $r_{t}$) and instantaneous components (inputs $u_{t}$).
The incorporated state space model provides sequential modulation on both residuals and instantaneous components.
Additional smoothed GMMs are applied to recursively estimate the priors of the unknown inputs.
The evaluation results show that SSVAE demonstrates advantages over other latent inference models when assessed using dynamics tracking and speech detection metrics.
Future work will focus on improving the estimation of the priors for the instantaneous components, and using the recursive estimation in an adversarial framework.   

\addtolength{\textheight}{-12cm}   








\bibliography{ssAutoencoder}
\bibliographystyle{plain}
\end{document}